%% file: main.tex
\documentclass[10pt,twocolumn,letterpaper]{article}

\usepackage[pagenumbers]{cvpr} 

\usepackage{multirow}
\usepackage{arydshln}
\usepackage{algorithmicx}
\usepackage{algpseudocode}
\usepackage{stackengine}
\usepackage{adjustbox}
\usepackage{hhline}
\usepackage{threeparttable}
\usepackage{booktabs}
\usepackage{soul}
\usepackage{fix-cm}

\DeclareMathOperator*{\argmax}{argmax}

\usepackage[linesnumbered,ruled,vlined]{algorithm2e}

\usepackage{color}
\usepackage{colortbl}
\usepackage{xcolor}
\usepackage{tikz}
\definecolor{gold}{HTML}{BD820B}
\definecolor{silver}{HTML}{909090}
\definecolor{bronze}{HTML}{9A5F26}
\definecolor{lgray}{gray}{0.95}

\definecolor{gaincolor}{RGB}{230, 245, 255}

\definecolor{Gray}{gray}{0.91}
\definecolor{LightCyan}{rgb}{0.82,0.82,1}
\newcolumntype{a}{>{\columncolor{Gray}}c}
\newcolumntype{B}{>{\columncolor{LightCyan}}c}

\usepackage{svg}
\svgsetup{inkscapelatex=true}

\usepackage{comment}

\usepackage{etoolbox}

\AtBeginEnvironment{algorithm}{%
  \setlength{\textfloatsep}{3pt}%
  \setlength{\intextsep}{3pt}%
  \setlength{\floatsep}{3pt}%
}

\input{preamble}

\definecolor{cvprblue}{rgb}{0.21,0.49,0.74}
\usepackage[pagebackref,breaklinks,colorlinks,allcolors=cvprblue]{hyperref}

\def\paperID{***}
\def\confName{CVPR}
\def\confYear{2025}

\title{
Rethinking Memory Design in SAM-Based Visual Object Tracking
}

\author{Mohamad Alansari, Muzammal Naseer, Hasan Al Marzouqi, Naoufel Werghi, and Sajid Javed \\
Department of Computer Science, Khalifa University, Abu Dhabi, United Arab Emirates\\
{\tt\small {100061914,muhammadmuzammal.naseer,hasan.almarzouqi,naoufel.werghi,sajid.javed}@ku.ac.ae}
}

\begin{document}
\maketitle
\input{sec/0_abstract}
\input{sec/1_intro}
\input{sec/2_related_work}

\input{sec/3_method}

\input{sec/4_results}

{
    \small
    \bibliographystyle{ieeenat_fullname}
    \bibliography{main}
}

\clearpage

\end{document}

%% file: preamble.tex
\usepackage{arydshln}
\usepackage{adjustbox}



%% file: sec/0_abstract.tex
\begin{abstract}
\noindent Memory has become the central mechanism enabling robust visual object tracking in modern segmentation-based frameworks.
Recent methods built upon Segment Anything Model 2 (SAM2) have demonstrated strong performance by refining how past observations are stored and reused.
However, existing approaches address memory limitations in a method-specific manner, leaving the broader design principles of memory in SAM-based tracking poorly understood.
Moreover, it remains unclear how these memory mechanisms transfer to stronger, next-generation foundation models such as Segment Anything Model 3 (SAM3).
In this work, we present a systematic memory-centric study of SAM-based visual object tracking.
We first analyze representative SAM2-based trackers and show that most methods primarily differ in how short-term memory frames are selected, while sharing a common object-centric representation.
Building on this insight, we faithfully reimplement these memory mechanisms within the SAM3 framework and conduct large-scale evaluations across ten diverse benchmarks, enabling a controlled analysis of memory design independent of backbone strength.
Guided by our empirical findings, we propose a unified hybrid memory framework that explicitly decomposes memory into short-term appearance memory and long-term distractor-resolving memory.
This decomposition enables the integration of existing memory policies in a modular and principled manner.
Extensive experiments demonstrate that the proposed framework consistently improves robustness under long-term occlusion, complex motion, and distractor-heavy scenarios on both SAM2 and SAM3 backbones.
Code is available at: \url{https://github.com/HamadYA/SAM3_Tracking_Zoo}.
\textbf{This is a preprint. Some results are being finalized and may be updated in a future revision.}
\end{abstract}

%% file: sec/1_intro.tex
\section{Introduction}
\label{sec:introduction}

Memory has emerged as a central design axis in modern visual object tracking (VOT) systems.
Recent memory-based frameworks~\cite{sam2,samurai,samite,him2sam,dam4sam,sam2long,xmem,cutie} have consistently achieved state-of-the-art (SOTA) performance on major VOT benchmarks~\cite{vot2020,vot2022,vots2023,vots2024}, substantially outperforming traditional template-based and search-and-match trackers.
A key driver behind this progress is the shift from rigid template matching to explicit memory mechanisms that preserve object identity and appearance over time.

The introduction of Segment Anything Model 2 (SAM2)~\cite{sam2} marked a decisive step in adapting large-scale segmentation foundation models to tracking by formulating VOT as a prompt-conditioned segmentation problem.
In this paradigm, temporal consistency is no longer enforced implicitly through handcrafted heuristics, but explicitly through a memory bank that conditions each frame on past observations.
As a result, memory evolves from a supporting component into the primary mechanism governing long-term tracking robustness.

Despite its success, directly deploying SAM2 for VOT reveals critical limitations.
SAM2 relies on a fixed First-In-First-Out (FIFO) memory update policy: one memory slot is permanently reserved for the initialization prompt, while the remaining slots are sequentially overwritten by recent frames.
Although simple and efficient, this strategy indiscriminately stores all frames regardless of prediction reliability, target visibility, or ambiguity.
Under long-term occlusion, fast motion, or visually similar distractors, unreliable frames are frequently written into memory, leading to error accumulation and identity drift.

These shortcomings have motivated a series of works that revisit SAM2’s memory design from different perspectives.
SAMURAI~\cite{samurai} introduces motion-aware memory filtering to suppress unreliable updates.
SAM2Long~\cite{sam2long} mitigates error propagation by maintaining multiple hypothesis pathways.
DAM4SAM~\cite{dam4sam} emphasizes distractor suppression through introspective memory updates.
SAMITE~\cite{samite} retains only calibrated prototype memories augmented with positional cues.
HiM2SAM~\cite{him2sam} separates short-term and long-term memory using hierarchical motion modeling.
While effective, these approaches are largely method-specific, addressing isolated failure modes with specialized mechanisms.
As a result, the broader design principles governing memory in SAM-based tracking remain fragmented and poorly understood.

More recently, Segment Anything Model 3 (SAM3)~\cite{sam3} further strengthens foundation models for video understanding.
Compared to SAM2, SAM3 employs stronger visual representations and a more conservative memory update strategy, retaining only frames where the target object is confidently present.
This reduces memory corruption and improves stability in many scenarios.
However, confidence-based retention alone is insufficient for resolving long-term occlusion, complex motion, and identity ambiguity, particularly in cluttered scenes.
Consequently, memory utilization in SAM3 remains suboptimal under severe temporal challenges, highlighting the need for more principled and adaptive memory designs that go beyond confidence filtering.

Taken together, these observations expose two fundamental and unresolved questions in SAM-based tracking:
\textbf{(1) How do the diverse memory mechanisms developed for SAM2 transfer to a stronger, next-generation architecture such as SAM3?}
\textbf{(2) Can the seemingly disparate memory strategies be unified under a common framework that better exploits their complementary strengths?}

In this work, we address these questions through a systematic memory-centric study of SAM-based trackers.
We first analyze representative SAM2-based methods to identify their core memory behaviors and failure modes.
We then faithfully reimplement these memory mechanisms within the SAM3 framework, enabling a controlled evaluation of how SAM2-era memory designs interact with a stronger foundation model.
Finally, guided by insights from both analyses, we propose a unified hybrid memory framework that decomposes memory into complementary components and integrates existing strategies in a principled manner.
Extensive experiments across diverse benchmarks demonstrate that our unified design consistently improves robustness over individual memory policies on both SAM2 and SAM3.

This paper makes the following contributions:

\begin{itemize}
    \item \textbf{A systematic memory-centric analysis of SAM-based tracking.}
    We provide a unified perspective on recent SAM2-based trackers by analyzing their memory mechanisms, revealing that most methods primarily differ in how short-term memory is selected and updated, while sharing a common object-centric representation.

    \item \textbf{A controlled transfer study from SAM2 to SAM3.}
    We faithfully reimplement representative SAM2-era memory mechanisms within the SAM3 framework and evaluate them across ten diverse benchmarks, offering the first large-scale analysis of how existing memory designs transfer to a stronger segmentation foundation model.

    \item \textbf{A unified hybrid memory framework.}
    We introduce a principled memory decomposition that separates short-term appearance memory from long-term distractor-resolving memory, enabling modular integration of existing memory policies.
    This framework unifies prior methods as special cases and yields consistent performance improvements over individual designs.

    \item \textbf{Extensive experimental validation.}
    We conduct comprehensive evaluations on challenging VOT benchmarks, demonstrating improved robustness under long-term occlusion, fast motion, and distractor-heavy scenarios on both SAM2 and SAM3 backbones.
\end{itemize}

%% file: sec/2_related_work.tex
\section{Related Work} \label{sec:relatedwork}

\subsection{Memory-Based Tracking Frameworks}
\noindent Memory has long been recognized as a critical component for VOT, enabling models to retain historical information and reason over long temporal horizons \cite{stm,aot,stcn}.
Early deep trackers primarily relied on implicit memory encoded in recurrent neural networks \cite{dcfst} or slowly updated appearance templates \cite{ostrack}.
While effective in short-term scenarios, these approaches struggled with long-term dependencies, severe occlusion, and target re-identification, as evidenced by analyses on tracking benchmarks \cite{vot2020,vot2022,vots2023,vots2024}.

To address these limitations, recent works have introduced explicit memory mechanisms that store historical visual information and retrieve it through similarity matching or attention-based aggregation.
Video object segmentation frameworks such as STM \cite{stm} pioneered the use of external memory banks that cache pixel-level features from previous frames and perform dense pixel-to-pixel matching for mask propagation.
Subsequent works, including STCN \cite{stcn}, further explored the design space of memory representation and retrieval by reformulating memory access as a spatio-temporal correspondence problem.
Building on this paradigm, XMem \cite{xmem} proposed a structured memory hierarchy that distinguishes between working, sensory, and long-term memory, enabling efficient retrieval and robust tracking over extended sequences.
XMem++ \cite{xmem++} further extends this framework toward production-level settings by improving robustness, scalability, and performance when only a few annotated frames are available.
CUTIE \cite{cutie} introduces object-level memory reading via object queries, which interact with pixel features through a query-based object transformer to improve robustness under distractors.

Despite their effectiveness, the aforementioned frameworks predominantly rely on dense pixel-level correspondence and explicit feature matching across frames, which incurs notable computational cost and assumes stable low-level appearance consistency.
Recent advances in segmentation foundation models motivate a different perspective on memory.
Rather than propagating masks through pixel-wise matching, SAM-based trackers formulate tracking as a prompt-conditioned segmentation problem, where memory acts as a high-level conditioning signal to preserve object identity over time.
This shift from bottom-up correspondence to top-down, concept-driven reasoning necessitates rethinking memory design, as exemplified by recent SAM2- and SAM3-based tracking frameworks discussed next.



\subsection{SAM-based Visual Object Tracking}
\noindent The introduction of SAM2 \cite{sam2} marked a significant departure from prior memory-based tracking frameworks by reformulating tracking as a prompt-conditioned segmentation problem.
Instead of relying on dense pixel-level correspondence, SAM2 maintains temporal consistency through an object-centric memory bank that stores past prompts and mask representations to condition future predictions.
This design enables efficient integration of large-scale segmentation foundation models into tracking pipelines, but also introduces new challenges in memory representation, update policies, and error accumulation.

Building upon SAM2, a series of works have proposed enhancements to its memory mechanism to improve robustness under challenging tracking conditions.
SAMURAI~\cite{samurai} introduces motion-aware memory selection to filter unreliable frames, reducing drift in crowded or fast-motion scenarios.
SAM2Long~\cite{sam2long} mitigates long-term error propagation by maintaining multiple segmentation hypotheses through a training-free multi-pathway inference strategy.
DAM4SAM~\cite{dam4sam} addresses distractor interference via introspective memory updates that jointly consider segmentation quality and tracking stability.
SAMITE~\cite{samite} further improves robustness by retaining only high-confidence memory prototypes and injecting explicit positional priors, while HiM2SAM~\cite{him2sam} separates short-term and long-term cues through a dual-memory structure with hierarchical motion modeling.

More recently, SAM3 \cite{sam3} advances this paradigm with stronger visual representations and a more conservative memory update strategy that retains only frames in which the target object is confidently present.
While this improves stability by reducing corrupted memory states, SAM3 largely inherits the object-centric, prompt-based memory formulation of SAM2.
As a result, fundamental questions remain regarding how existing SAM2-era memory policies transfer to this stronger architecture and whether their complementary logic can be unified to form more effective memory mechanisms.

%% file: sec/3_method.tex
\section{Method} \label{sec:method}
\noindent 

\subsection{Overview}
\noindent Our work studies memory mechanisms in SAM-based visual object tracking and proposes hybrid memory designs that improve robustness across model generations.
We make two key observations: (i) existing SAM2-based trackers primarily differ in how short-term memory is selected and updated, and (ii) DAM4SAM provides a modular decomposition of memory that naturally supports replacing individual components.
Based on these observations, our method consists of two stages.
First, we faithfully recreate representative SAM2-based trackers under the SAM3 backbone to evaluate how their memory logic transfers to a stronger foundation model.
Second, we construct hybrid memory designs by replacing the Recent Appearance Memory (RAM) component in DAM4SAM with alternative memory policies, and evaluate these hybrids under both SAM2 and SAM3.

\subsection{Recreating SAM2 Trackers in SAM3} \label{sec:sam3_transfer}
\noindent Let $\Phi^{(2)}$ and $\Phi^{(3)}$ denote the frozen SAM2 and SAM3 tracking backbones, respectively.
Given a video sequence $\{I_t\}_{t=1}^{T}$ and an initialization prompt, SAM-based tracking predicts a segmentation mask $\hat{M}_t$ at each time step conditioned on a memory bank $B_t$.
To analyze transferability, we reimplement SAM2-based trackers by preserving their memory logic while replacing $\Phi^{(2)}$ with $\Phi^{(3)}$.
Specifically, we keep the original proposal selection rules, memory update criteria, and memory layout for each method.
Only the image encoder, prompt encoder, and mask decoder are replaced by their SAM3 counterparts.
All model parameters remain frozen, and hyperparameters are matched to the original implementations whenever possible.
This controlled protocol isolates the effect of improved representations on memory behavior.

\subsection{Memory Mechanisms and Hybridization}
\noindent We consider single-object VOT on a video sequence
$\{I_t\}_{t=1}^{T}$, where the target is specified in the first frame by an initialization prompt $m_{\text{init}}$ (mask or bounding box).
At each time step $t$, a SAM-based tracker predicts a segmentation mask
$\hat{M}_t$ for the current frame $I_t$ conditioned on an explicit memory bank $B_t$.

\noindent \textbf{SAM2 Memory Update.}
In SAM2, the memory bank is object-centric and of fixed capacity.
One slot is permanently reserved for the initialization prompt, which serves as a stable anchor throughout tracking.
The remaining $K = N_{\text{mem}} - 1$ slots store recent frame memories encoded from past predictions.
Formally, the memory bank at time $t$ is:
\begin{equation}
B_t^{\text{SAM2}} = \{ m_{\text{init}} \} \cup \{ m_{t_1}, m_{t_2}, \dots, m_{t_K} \},
\end{equation}
where each memory entry $m_{t_i}$ is generated by the memory encoder from $(I_{t_i}, \hat{M}_{t_i})$.

SAM2 updates its dynamic memory using a First-In-First-Out (FIFO) policy.
At each frame, the newly generated memory entry $m_t$ is inserted into the memory bank, and the oldest dynamic entry is discarded.
This update strategy does not explicitly consider prediction confidence, target visibility, or ambiguity.
Consequently, frames with inaccurate segmentation, severe occlusion, or distractor confusion may be written into memory and subsequently reinforced through memory attention, leading to error accumulation and identity drift in challenging scenarios.

\noindent \textbf{DAM4SAM: RAM and DRM Decomposition.}
DAM4SAM decomposes the dynamic memory of SAM-based trackers into two complementary components: Recent Appearance Memory (RAM) and Distractor Resolving Memory (DRM).
The overall memory bank is defined as \(B_t^{\text{DAM}} = \{ m_{\text{init}} \} \cup B_t^{\text{RAM}} \cup B_t^{\text{DRM}}\), where RAM captures short-term target appearance for accurate segmentation, and DRM stores a small set of long-term anchor frames that encode target--distractor relationships and support re-detection.

RAM is designed to retain recent, reliable target appearances.
Instead of updating at every frame, RAM is updated sparsely to reduce redundancy and prevent contamination.
Frames are admitted into RAM only when the target is predicted as present and when a minimum temporal gap since the last RAM update is satisfied.
Within RAM, entries are maintained using a FIFO policy over a fixed capacity $K_{\text{RAM}}$.

DRM is responsible for robustness against distractors and long-term ambiguities.
It stores a small set of anchor frames selected based on proposal-level disagreement in SAM’s multi-mask predictions, indicating the presence of visually similar distractors.
To ensure stability, additional quality constraints on prediction confidence and mask area consistency are imposed before committing a frame as a DRM anchor.
DRM entries are updated sparsely in time and retained for long durations, providing strong cues for re-detection and disambiguation.

\noindent \textbf{Hybrid Memory: Replacing RAM.}
A key observation is that most SAM2-based tracking methods primarily differ in how short-term memory frames are selected and updated, while sharing a common object-centric memory representation.
DAM4SAM’s explicit decomposition of memory into RAM and DRM naturally enables modular replacement of the RAM component.

We construct hybrid memory mechanisms by retaining the DRM component of DAM4SAM unchanged and replacing the RAM update policy with alternative memory selection strategies from the literature \cite{samurai,sam2long,samite,him2sam}.
Formally, the hybrid memory bank at time $t$ is defined as:
\begin{equation}
B_t^{\text{Hybrid}} = \{ m_{\text{init}} \} \cup B_t^{\text{RAM}\leftarrow \pi} \cup B_t^{\text{DRM}},
\end{equation}
where $B_t^{\text{RAM}\leftarrow \pi}$ denotes the RAM constructed using policy $\pi$.

Across all hybrids, the RAM capacity $K_{\text{RAM}}$ and the DRM configuration are kept identical to DAM4SAM to ensure fair comparison.
We keep DRM unchanged across all hybrids because it encodes long-term, distractor-critical states whose semantics are orthogonal to short-term appearance selection.
This allows us to isolate the effect of short-term memory policies while preserving consistent long-term disambiguation.
Only the logic governing RAM selection is replaced.
Using the above framework, we instantiate four hybrid trackers by integrating representative SAM2-based memory policies into the RAM component.
All hybrid trackers are implemented in a training-free manner, modifying only memory selection logic while keeping all model parameters frozen.

\noindent \textbf{(a) SAMURAI-DRM.}
In this hybrid, we adopt SAMURAI’s motion-aware scoring as the RAM admission policy.
At frame $t$, the SAM mask decoder outputs a set of candidate masks
$\{\mathcal{M}_{t,i}\}$ with affinity scores $s_{\text{mask},t,i}$ and object scores $s_{\text{obj},t,i}$.
A Kalman filter is used to predict the target bounding box $\hat{b}_{t|t-1}$ based on past motion.
For each candidate mask, a motion-consistency score is computed as:
\begin{equation}
s_{\text{kf},t,i} = \operatorname{IoU}(\hat{b}_{t|t-1}, b_{t,i}),
\end{equation}
where $b_{t,i}$ is the bounding box of $\mathcal{M}_{t,i}$.
The final output mask is selected by maximizing a weighted combination of motion and affinity scores:
\begin{equation}
\mathcal{M}_t^* =
\argmax_{\mathcal{M}_{t,i}}
\big(
\alpha\, s_{\text{kf},t,i} + (1-\alpha)\, s_{\text{mask},t,i}
\big),
\quad \text{s.t. } s_{\text{obj},t,i}>0 .
\end{equation}

For memory construction, a frame $i$ is admitted into RAM only if:
\begin{equation}
s_{\text{mask},i} \ge \tau_{\text{mask}}, \quad
s_{\text{obj},i} \ge \tau_{\text{obj}}, \quad
s_{\text{kf},i} \ge \tau_{\text{kf}} .
\end{equation}
Starting from the current frame, we select the most recent $K_{\text{RAM}}$ frames satisfying these conditions to form the motion-consistent RAM.

\noindent \textbf{(b) SAM2Long-DRM.}
In this hybrid, we adopt SAM2Long’s multi-hypothesis selection strategy as the RAM admission policy.
At each time step $t$, we maintain $P$ memory pathways, each associated with a memory bank $B_t^p$ and a cumulative score $S_p[t]$.
Conditioned on $B_t^p$, the SAM mask decoder generates three candidate masks with predicted IoU scores $\{\text{IoU}_t^{p,k}\}_{k=1}^3$.
Each pathway is expanded and scored as:
\begin{equation}
S_{p,k}[t] = S_p[t-1] + \log(\text{IoU}_t^{p,k} + \epsilon),
\end{equation}
after which only the top-$P$ pathways are retained.

For RAM construction, only frames belonging to the highest-scoring pathway are eligible for memory insertion.
A frame $i$ is admitted into RAM if:
\begin{equation}
\text{IoU}_i \ge \tau_{\text{IoU}} \quad \text{and} \quad o_i > 0,
\end{equation}
where $o_i$ denotes the predicted object presence score.
Starting from the current frame, we select the most recent $K_{\text{RAM}}$ frames satisfying these conditions to form the short-term memory.
This strategy replaces FIFO-based RAM updates with uncertainty-aware, trajectory-consistent memory selection.

\noindent \textbf{(c) SAMITE-DRM.}
In this hybrid, we adopt SAMITE’s prototypical memory selection strategy as the RAM admission policy.
For each frame $t$, SAMITE extracts a foreground prototype $P_t^{FG}$ and background prototype $P_t^{BG}$ from the predicted mask.
To construct RAM, candidate frames within a recent temporal window $\mathcal{C}=\{t-m,\dots,t-2\}$ are scored based on their consistency with two anchors: the first frame and the previous frame.
Specifically, for each candidate frame $\tau \in \mathcal{C}$, a calibration score is computed as
\begin{equation}
S_\tau = (1-\alpha)\,\text{Cos}(P_\tau^{FG}, P_1^{FG})
+ \alpha\,\text{Cos}(P_\tau^{FG}, P_{t-1}^{FG}),
\end{equation}
where $P_1^{FG}$ and $P_{t-1}^{FG}$ serve as feature-wise and position-wise anchors, respectively.

The RAM is formed by including the first and previous frames as anchors, and selecting the remaining $K_{\text{RAM}}-2$ frames with the highest calibration scores:
\begin{equation}
B_t^{\text{RAM}} = \text{Top-}K_{\text{RAM}}(\{m_\tau\}_{\tau\in\mathcal{C}}, S_\tau).
\end{equation}
This replaces FIFO-based RAM updates with prototype-calibrated short-term memory that favors accurate and spatially nearby target appearances.

\noindent \textbf{(d) HiM2SAM-DRM.}
In this hybrid, we use HiM2SAM’s motion-enhanced confidence filtering as the RAM admission policy.
At frame $t$, SAM produces multiple mask proposals with confidence $s_{\text{iou},t,i}$.
HiM2SAM computes a motion-aware confidence score by combining segmentation confidence with motion consistency.
Using a coarse motion estimator (e.g., Kalman filter), the confidence is
\begin{equation}
s_{\text{conf},t,i} = \alpha\, s_{\text{coarse},t,i} + (1-\alpha)\, s_{\text{iou},t,i}.
\end{equation}
If the maximum confidence is below a threshold $\tau$, a fine motion estimator is activated, and the confidence is refined as
\begin{equation}
s_{\text{conf},t,i} = \alpha\, s_{\text{coarse},t,i} + \beta\, s_{\text{fine},t,i} + (1-\alpha-\beta)\, s_{\text{iou},t,i}.
\end{equation}

For RAM construction, a frame $i$ is admitted into short-term memory only if its selected prediction is high-confidence:
\begin{equation}
s_{\text{conf},i} \ge \tau_{\text{mem}} \quad \text{and} \quad \hat{M}_i \neq \emptyset,
\end{equation}
after which RAM retains the most recent $K_{\text{RAM}}$ accepted frames (FIFO).
This yields a hybrid tracker in which short-term memory is motion-consistent and selectively refreshed, while long-term disambiguation is provided by DAM4SAM’s DRM.

%% file: sec/4_results.tex
\begin{table*}[!h]
\centering

\caption{
Comparison of SAM2-based tracking baselines on five datasets \cite{lasot,lasotext,tnl2k,got10k,trackingnet}.
SR-DRM denotes SAMURAI with DAM4SAM, SA-DRM denotes SAMITE with DAM4SAM, and HiM-DRM denotes HiM2SAM with DAM4SAM.
}

\vspace{-1em}

\setlength{\tabcolsep}{1pt}
		\scalebox{0.8}[0.8]{

\begin{tabular}{l|ccc|ccc|ccc|ccc|ccc}

\hline

\rowcolor{Gray} 
Method
& \multicolumn{3}{c|}{LaSOT \cite{lasot}}
& \multicolumn{3}{c|}{LaSOT\textsubscript{ext} \cite{lasotext}}
& \multicolumn{3}{c|}{TNL2K \cite{tnl2k}}
& \multicolumn{3}{c|}{GOT-10k \cite{got10k}}
& \multicolumn{3}{c}{TrackingNet \cite{trackingnet}}
\\

\cline{2-16}
\rowcolor{Gray}
& S & NP & P 
& S & NP & P 
& S & NP & P 
& AO & SR\textsubscript{0.50} & SR\textsubscript{0.75} 
& S & NP & P 
\\

\hline 

SAM2-T \cite{sam2} 
& 67.0 & 74.0 & 71.5 
& 50.8 & 62.4 & 58.9 
& 55.8 & 74.2 & 60.7 
& 78.5 & 88.9 & 72.4 
& 83.9 & 89.8 & 85.9 
\\

SAM2-S \cite{sam2} 
& 66.3 & 73.4 & 71.0 
& 54.9 & 68.2 & 64.6 
& 56.4 & 75.6 & 61.9 
& 78.3 & 87.9 & 73.0 
& 84.4 & 90.4 & 86.9 
\\

SAM2-B \cite{sam2} 
& 66.2 & 73.8 & 71.2 
& 54.3 & 67.7 & 63.6 
& 55.4 & 74.1 & 60.7 
& 78.0 & 88.8 & 71.6 
& 84.6 & 90.5 & 87.2 
\\

SAM2-L \cite{sam2} 
& 68.5 & 76.1 & 73.6 
& 56.8 & 71.1 & 67.0 
& 56.7 & 75.4 & 62.5 
& 80.8 & 91.3 & 75.5 
& 85.3 & 91.3 & 88.2 
\\

\hline

DAM4SAM-T \cite{dam4sam} 
& 72.4 & 80.2 & 77.7 
& 57.4 & 70.6 & 67.6 
& 59.4 & 79.0 & 65.5 
& 78.3 & 89.3 & 72.3 
& 83.9 & 89.7 & 85.3 
\\

DAM4SAM-S \cite{dam4sam} 
& 72.6 & 80.5 & 78.1 
& 59.2 & 73.2 & 70.2 
& 59.4 & 79.3 & 65.8 
& 74.0 & 82.0 & 66.4 
& 77.2 & 79.5 & 74.8 
\\

DAM4SAM-B \cite{dam4sam} 
& 73.2 & 81.3 & 79.0 
& 57.1 & 70.6 & 66.5 
& 58.4 & 78.1 & 64.6 
& 78.9 & 89.3 & 72.3 
& 84.7 & 90.5 & 86.7 
\\

DAM4SAM-L \cite{dam4sam} 
& 75.1 & 83.3 & 81.1 
& 60.9 & 75.3 & 72.2 
& 59.8 & 79.8 & 66.8 
& 81.1 & 91.4 & 77.2 
& 85.3 & 90.9 & 87.4 
\\

\hline

SAMURAI-T \cite{samurai} 
& 69.3 & 76.4 & 73.8 
& 53.6 & 65.9 & 62.2 
& 50.6 & 68.2 & 53.3 
& 77.5 & 87.8 & 71.0 
& 77.7 & 83.5 & 79.0 
\\

SAMURAI-S \cite{samurai} 
& 70.0 & 77.6 & 75.2 
& 56.5 & 69.8 & 66.3 
& 45.7 & 61.5 & 46.7 
& 76.7 & 86.4 & 71.0 
& 77.9 & 83.4 & 79.2 
\\

SAMURAI-B \cite{samurai} 
& 70.7 & 78.7 & 76.2 
& 55.9 & 69.5 & 65.6 
& 47.8 & 64.3 & 50.1 
& 77.8 & 88.6 & 71.1 
& 79.3 & 84.9 & 81.2 
\\

SAMURAI-L \cite{samurai} 
& 74.2 & 82.7 & 80.2 
& 61.0 & 73.9 & 72.2 
& 50.6 & 67.5 & 54.2 
& 81.7 & 92.2 & 76.9 
& 85.3 & 88.2 & 85.0 
\\

\hline

SR-DRM-T
& 73.1 \tiny{\textcolor{gray}{\textbf{(0.7$\uparrow$)}}} & 81.1 \tiny{\textcolor{gray}{\textbf{(0.9$\uparrow$})}} & 78.1 \tiny{\textcolor{gray}{\textbf{(0.4$\uparrow$)}}} 
& 58.9 \tiny{\textcolor{gray}{\textbf{(1.5$\uparrow$)}}} & 72.6 \tiny{\textcolor{gray}{\textbf{(2.0$\uparrow$)}}} & 69.4 \tiny{\textcolor{gray}{\textbf{(1.8$\uparrow$)}}} 
& 59.4 & 78.5 & 64.6 
& - & - & - 
& 84.4 \tiny{\textcolor{gray}{\textbf{(0.5$\uparrow$)}}} & 90.0 \tiny{\textcolor{gray}{\textbf{(0.3$\uparrow$)}}} & 85.8 \tiny{\textcolor{gray}{\textbf{(0.5$\uparrow$)}}} 
\\

SR-DRM-S
& 73.8 \tiny{\textcolor{gray}{\textbf{(1.2$\uparrow$)}}} & 81.8 \tiny{\textcolor{gray}{\textbf{(1.3$\uparrow$)}}} & 79.6 \tiny{\textcolor{gray}{\textbf{(1.5$\uparrow$)}}} 
& 60.0 \tiny{\textcolor{gray}{\textbf{(0.8$\uparrow$)}}} & 73.9 \tiny{\textcolor{gray}{\textbf{(0.7$\uparrow$)}}} & 70.7 \tiny{\textcolor{gray}{\textbf{(7.5$\uparrow$)}}} 
& 59.3 & 78.9 & 65.2 
& - & - & - 
& 84.9 \tiny{\textcolor{gray}{\textbf{(7.7$\uparrow$)}}} & 90.5 \tiny{\textcolor{gray}{\textbf{(11.0$\uparrow$)}}} & 86.6 \tiny{\textcolor{gray}{\textbf{(11.8$\uparrow$)}}} 
\\

SR-DRM-B
& 74.5 \tiny{\textcolor{gray}{\textbf{(1.3$\uparrow$)}}} & 82.8 \tiny{\textcolor{gray}{\textbf{(1.5$\uparrow$)}}} & 80.4 \tiny{\textcolor{gray}{\textbf{(1.4$\uparrow$)}}} 
& 58.0 \tiny{\textcolor{gray}{\textbf{(0.9$\uparrow$)}}} & 71.7 \tiny{\textcolor{gray}{\textbf{(1.1$\uparrow$)}}} & 67.9 \tiny{\textcolor{gray}{\textbf{(1.4$\uparrow$)}}} 
& 59.8 \tiny{\textcolor{gray}{\textbf{(1.4$\uparrow$)}}} & 79.2 \tiny{\textcolor{gray}{\textbf{(1.1$\uparrow$)}}} & 65.9 \tiny{\textcolor{gray}{\textbf{(1.3$\uparrow$)}}} 
& - & - & - 
& 85.1 \tiny{\textcolor{gray}{\textbf{(0.4$\uparrow$)}}} & 90.7 \tiny{\textcolor{gray}{\textbf{(0.2$\uparrow$)}}} & 87.2 \tiny{\textcolor{gray}{\textbf{(0.5$\uparrow$)}}} 
\\

SR-DRM-L
& 74.9 & 83.1 & 80.6 
& 61.1 \tiny{\textcolor{gray}{\textbf{(0.2$\uparrow$)}}} & 75.6 \tiny{\textcolor{gray}{\textbf{(0.3$\uparrow$)}}} & 71.9 \tiny{\textcolor{gray}{\textbf{(0.3$\downarrow$)}}} 
& 60.4 \tiny{\textcolor{gray}{\textbf{(0.6$\uparrow$)}}} & 80.3 \tiny{\textcolor{gray}{\textbf{(0.5$\uparrow$)}}} & 66.9 \tiny{\textcolor{gray}{\textbf{(0.1$\uparrow$)}}} 
& - & - & - 
& 85.4 \tiny{\textcolor{gray}{\textbf{(0.1$\uparrow$)}}} & 90.9 & 87.5 \tiny{\textcolor{gray}{\textbf{(0.1$\uparrow$)}}} 
\\

\hline

SAMITE-T \cite{samite}
& 72.8 & 80.6 & 78.3 
& 56.0 & 68.2 & 64.8 
& 60.4 & 79.6 & 66.3 
& 79.4 & 89.8 & 73.4 
& 84.4 & 89.9 & 85.6 
\\

SAMITE-S \cite{samite}
& 73.0 & 81.3 & 79.2 
& 58.3 & 71.9 & 68.6 
& 61.0 & 80.7 & 67.3 
& 79.4 & 89.2 & 73.6 
& 85.1 & 90.4 & 86.7 
\\

SAMITE-B \cite{samite}
& 74.9 & 83.4 & 81.4 
& 60.7 & 73.1 & 71.2 
& 60.7 & 80.2 & 67.2 
& 78.9 & 89.9 & 72.5 
& 85.2 & 90.5 & 86.9 
\\

SAMITE-L \cite{samite}
& 74.7 & 83.3 & 81.1 
& 60.5 & 75.4 & 72.0 
& 61.7 & 81.3 & 68.2 
& 80.3 & 90.1 & 75.4 
& 85.7 & 91.1 & 87.9 
\\

\hline

SA-DRM-T
& - & - & - 
& - & - & - 
& - & - & - 
& - & - & - 
& - & - & - 
\\

SA-DRM-S
& - & - & - 
& - & - & - 
& - & - & - 
& - & - & - 
& - & - & - 
\\

SA-DRM-B
& - & - & - 
& - & - & - 
& - & - & - 
& - & - & - 
& - & - & - 
\\

SA-DRM-L
& - & - & - 
& - & - & - 
& - & - & - 
& - & - & - 
& - & - & - 
\\

\hline

HiM2SAM-T \cite{him2sam}
& 72.4 & 80.3 & 78.0 
& 57.0 & 69.5 & 66.5 
& 59.9 & 78.6 & 65.3 
& 76.7 & 87.3 & 70.6 
& 84.7 & 89.8 & 86.0 
\\

HiM2SAM-S \cite{him2sam}
& 73.1 & 81.0 & 78.8 
& 58.7 & 72.1 & 68.9 
& 59.7 & 78.9 & 65.6 
& 74.3 & 83.5 & 69.1 
& 85.0 & 90.2 & 86.6 
\\

HiM2SAM-B \cite{him2sam}
& 73.4 & 81.7 & 79.5 
& 57.7 & 71.4 & 67.9 
& 60.6 & 79.9 & 67.0 
& 74.8 & 85.2 & 68.6 
& 85.4 & 90.8 & 87.3 
\\

HiM2SAM-L \cite{him2sam}
& 75.1 & 83.2 & 81.0 
& 61.3 & 75.7 & 72.8 
& 60.4 & 79.5 & 66.7 
& 74.2 & 83.5 & 69.3 
& 85.8 & 91.0 & 87.9 
\\

\hline

HiM-DRM-T
& - & - & - 
& - & - & - 
& - & - & - 
& 78.7 \tiny{\textcolor{gray}{\textbf{(0.4$\uparrow$)}}} & 88.9 & 72.7 \tiny{\textcolor{gray}{\textbf{(0.4$\uparrow$)}}} 
& - & - & - 
\\

HiM-DRM-S
& - & - & - 
& - & - & - 
& - & - & - 
& 79.4 \tiny{\textcolor{gray}{\textbf{(5.4$\uparrow$)}}} & 88.6 \tiny{\textcolor{gray}{\textbf{(6.6$\uparrow$)}}} & 74.2 \tiny{\textcolor{gray}{\textbf{(7.8$\uparrow$)}}} 
& - & - & - 
\\

HiM-DRM-B
& - & - & - 
& - & - & - 
& - & - & - 
& 79.8 \tiny{\textcolor{gray}{\textbf{(0.9$\uparrow$)}}} & 89.2 & 74.2 \tiny{\textcolor{gray}{\textbf{(1.9$\uparrow$)}}} 
& - & - & = 
\\

HiM-DRM-L
& - & - & - 
& - & - & - 
& - & - & - 
& 80.4 & 89.8 & 76.2 
& - & - & - 
\\

\hline

\end{tabular}
}
\label{tab:sam2_bbox_results}
\vspace{-1em}
\end{table*}

\begin{table*}[!h]
\centering

\caption{
Comparison of SAM3-based tracking baselines on five datasets \cite{lasot,lasotext,tnl2k,got10k,trackingnet}.
SR-DRM-3 denotes SAMURAI-3 with DAM4SAM-3, SA-DRM-3 denotes SAMITE-3 with DAM4SAM-3, and HiM-DRM-3 denotes HiM2SAM-3 with DAM4SAM-3.
}

\vspace{-1em}

\setlength{\tabcolsep}{5pt}
		\scalebox{0.9}[0.9]{

\begin{tabular}{l|ccc|ccc|ccc|ccc|ccc}

\hline

\rowcolor{Gray} 
Method
& \multicolumn{3}{c|}{LaSOT \cite{lasot}}
& \multicolumn{3}{c|}{LaSOT\textsubscript{ext} \cite{lasotext}}
& \multicolumn{3}{c|}{TNL2K \cite{tnl2k}}
& \multicolumn{3}{c|}{GOT-10k \cite{got10k}}
& \multicolumn{3}{c}{TrackingNet \cite{trackingnet}}
\\

\cline{2-16}
\rowcolor{Gray}
& S & NP & P 
& S & NP & P 
& S & NP & P 
& AO & SR\textsubscript{0.50} & SR\textsubscript{0.75} 
& S & NP & P 
\\

\hline 

SAM3 \cite{sam3}
& 74.7 & 82.0 & 78.8 
& 63.8 & 78.5 & 75.5 
& 64.2 & 84.4 & 71.9 
& 84.5 & 94.6 & 82.2 
& 86.6 & 91.2 & 88.2 
\\

SAM3Long
& 75.2 & 82.3 & 79.6 
& 62.7 & 77.4 & 73.9 
& 64.6 & 84.8 & 72.6 
& 84.7 & 95.2 & 82.4 
& 86.7 & 91.4 & 88.4 
\\

DAM4SAM-3
& 76.5 & 83.9 & 80.9 
& 66.0 & 81.3 & 78.6 
& 65.6 & 86.0 & 73.7 
& 84.0 & 94.2 & 81.5 
& 86.8 & 91.5 & 88.5 
\\

SAMURAI-3
& 75.2 & 82.3 & 79.4 
& 62.6 & 77.1 & 74.3 
& 64.9 & 85.0 & 72.7 
& 84.4 & 94.4 & 81.9 
& 86.5 & 91.1 & 88.2 
\\

SR-DRM-3
& -- & -- & -- 
& -- & -- & -- 
& -- & -- & -- 
& -- & -- & -- 
& -- & -- & -- 
\\

SAMITE-3
& 77.4 & 85.1 & 82.4 
& 65.3 & 80.2 & 77.3 
& 65.1 & 85.4 & 72.8 
& 83.9 & 94.2 & 81.2 
& 86.7 & 91.5 & 88.6 
\\

SA-DRM-3
& -- & -- & -- 
& -- & -- & -- 
& -- & -- & -- 
& -- & -- & -- 
& -- & -- & -- 
\\

HiM2SAM-3
& 75.6 & 82.8 & 79.6 
& 62.9 & 77.1 & 74.0 
& 64.8 & 84.9 & 72.5 
& 84.1 & 94.1 & 81.8 
& 86.6 & 91.2 & 88.2 
\\

HiM-DRM-3
& -- & -- & -- 
& -- & -- & -- 
& -- & -- & -- 
& -- & -- & -- 
& -- & -- & -- 
\\

\hline

\end{tabular}
}
\label{tab:sam3_bbox_results}
\vspace{-1em}
\end{table*}

\begin{table*}[h!]
\centering

\caption{
Comparison of SAM2-based tracking baselines on VOT20-24 \cite{vot2020, vot2022, vots2024}, and DiDi \cite{dam4sam} datasets.
SR-DRM denotes SAMURAI with DAM4SAM, SA-DRM denotes SAMITE with DAM4SAM, and HiM-DRM denotes HiM2SAM with DAM4SAM.
}

\label{tab:sam2_vot}

\vspace{-1em}
\setlength{\tabcolsep}{8pt}
		\scalebox{0.9}[0.9]{
        

\begin{tabular}{l|ccc|ccc|ccc|ccc}

\hline

\rowcolor{Gray} Method & \multicolumn{3}{c}{\textbf{VOT20}} & \multicolumn{3}{c}{\textbf{VOT22}} & \multicolumn{3}{c}{\textbf{VOTS24}} & \multicolumn{3}{c}{\textbf{DiDi}} \\

\rowcolor{Gray} & \textit{Q} & \textit{Acc} & \textit{Rob} & \textit{Q} & \textit{Acc} & \textit{Rob} & \textit{Q} & \textit{Acc} & \textit{Rob} & \textit{Q} & \textit{Acc} & \textit{Rob} \\

\hline

SAM2-T \cite{sam2}
& 62.1 & 72.3 & 93.2
& 62.3 & 72.4 & 93.4 
& 62.8 & 75.6 & 77.9 
& 60.5 & 70.2 & 84.9
\\

SAM2-S \cite{sam2}
& 64.3 & 73.3 & 93.5 
& 64.4 & 73.3 & 93.6 
& 60.5 & 75.1 & 76.7
& 62.4 & 70.9 & 86.7
\\

SAM2-B \cite{sam2}
& 64.4 & 73.8 & 93.4 
& 64.0 & 73.7 & 93.2 
& 62.8 & 77.0 & 76.2
& 63.6 & 72.0 & 87.0
\\

SAM2-L \cite{sam2}
& 68.1 & 77.8 & 94.1
& 69.2 & 77.9 & 94.6 
& 66.1 & 79.1 & 79.0
& 64.7 & 72.3 & 87.7
\\

\hline

DAM4SAM-T \cite{dam4sam}
& 67.0 & 73.7 & 96.1 
& 68.1 & 73.9 & 96.4
& 65.3 & 75.3 & 84.2
& 63.2 & 69.3 & 89.8
\\

DAM4SAM-S \cite{dam4sam}
& 54.9 & 73.3 & 87.3 
& 55.6 & 73.2 & 88.4
& 59.3 & 65.6 & 84.9
& 66.3 & 71.0 & 92.3
\\

DAM4SAM-B \cite{dam4sam}
& 70.1 & 74.5 & 96.5 
& 70.4 & 74.4 & 96.4
& 68.3 & 77.0 & 86.5
& 66.4 & 70.8 & 93.1
\\

DAM4SAM-L \cite{dam4sam}
& 72.3 & 79.6 & 96.1
& 75.0 & 79.7 & 97.1
& 71.1 & 79.3 & 86.4
& 69.4 & 72.7 & 94.4
\\

\hline

SAMURAI-T \cite{samurai}
& 63.3 & 72.5 & 93.2 
& 63.7 & 73.0 & 93.9
& 63.9 & 75.1 & 83.7
& 62.7 & 68.1 & 89.9
\\

SAMURAI-S \cite{samurai}
& 65.1 & 71.2 & 94.2
& 65.3 & 72.4 & 94.9
& 64.5 & 75.2 & 83.3
& 64.6 & 70.1 & 90.6
\\

SAMURAI-B \cite{samurai}
& 66.4 & 73.2 & 95.0 
& 66.8 & 73.8 & 95.3 
& 64.6 & 75.5 & 83.3
& 67.1 & 71.5 & 93.0
\\

SAMURAI-L \cite{samurai}
& 67.4 & 74.1 & 94.2
& 69.3 & 74.4 & 95.1
& 67.3 & 77.6 & 85.1
& 67.0 & 71.4 & 92.3
\\

\hline

SR-DRM-T
& - & - & - 
& - & - & - 
& - & - & - 
& - & - & - 
\\

SR-DRM-S
& - & - & - 
& - & - & - 
& - & - & - 
& - & - & - 
\\

SR-DRM-B
& - & - & - 
& - & - & - 
& - & - & - 
& - & - & - 
\\

SR-DRM-L
& - & - & - 
& - & - & - 
& - & - & - 
& - & - & - 
\\

\hline

SAMITE-T \cite{samite}
& 66.4 & 73.5 & 96.3 
& 67.5 & 73.2 & 96.4 
& 64.1 & 74.3 & 83.4 
& 65.7 & 70.4 & 91.9 
\\

SAMITE-S \cite{samite}
& 66.4 & 74.6 & 95.9 
& 67.4 & 74.5 & 96.0 
& 66.2 & 76.1 & 85.3 
& 66.4 & 71.1 & 91.8 
\\

SAMITE-B \cite{samite}
& 69.4 & 75.9 & 96.2 
& 69.4 & 75.9 & 96.1 
& 67.7 & 76.8 & 86.0 
& 69.4 & 72.8 & 94.8 
\\

SAMITE-L \cite{samite}
& 70.4 & 78.6 & 95.1 
& 72.5 & 78.5 & 95.8 
& 68.6 & 77.5 & 86.9 
& 68.4 & 72.3 & 93.6 
\\

\hline

SA-DRM-T
& - & - & - 
& - & - & - 
& - & - & - 
& - & - & - 
\\

SA-DRM-S
& - & - & - 
& - & - & - 
& - & - & - 
& - & - & - 
\\

SA-DRM-B
& - & - & - 
& - & - & - 
& - & - & - 
& - & - & - 
\\

SA-DRM-L
& - & - & - 
& - & - & - 
& - & - & - 
& - & - & - 
\\

\hline

HiM2SAM-T \cite{him2sam}
& 60.7 & 73.5 & 93.3 
& 62.4 & 73.7 & 94.3 
& 66.0 & 75.1 & 85.0 
& 63.3 & 68.7 & 89.9 
\\

HiM2SAM-S \cite{him2sam}
& 61.4 & 74.9 & 92.8 
& 63.1 & 75.1 & 93.8 
& 65.2 & 74.8 & 83.1 
& 63.3 & 69.1 & 89.9 
\\

HiM2SAM-B \cite{him2sam}
& 61.0 & 75.0 & 91.7 
& 63.5 & 75.1 & 93.8 
& 67.4 & 76.7 & 86.6 
& 68.1 & 71.1 & 94.6 
\\

HiM2SAM-L \cite{him2sam}
& 61.6 & 79.1 & 90.1 
& 61.3 & 78.5 & 90.0 
& 66.2 & 76.1 & 84.2 
& 67.0 & 71.2 & 91.9 
\\

\hline

HiM-DRM-T
& - & - & - 
& - & - & - 
& - & - & - 
& - & - & - 
\\

HiM-DRM-S
& - & - & - 
& - & - & - 
& - & - & - 
& - & - & - 
\\

HiM-DRM-B
& - & - & - 
& - & - & - 
& - & - & - 
& - & - & - 
\\

HiM-DRM-L
& - & - & - 
& - & - & - 
& - & - & - 
& - & - & - 
\\

\hline

\end{tabular}
}
\end{table*}

\begin{table*}[h!]
\centering
\caption{
Comparison of SAM2-based tracking baselines on VOT20-24 \cite{vot2020, vot2022, vots2024}, and DiDi \cite{dam4sam} datasets.
SR-DRM-3 denotes SAMURAI-3 with DAM4SAM-3, SA-DRM-3 denotes SAMITE-3 with DAM4SAM-3, and HiM-DRM-3 denotes HiM2SAM-3 with DAM4SAM-3.
}

\label{tab:sam3_vot}

\vspace{-1em}
\setlength{\tabcolsep}{8pt}
		\scalebox{0.9}[0.9]{
        

\begin{tabular}{l|ccc|ccc|ccc|ccc}

\hline

\rowcolor{Gray} Method & \multicolumn{3}{c}{\textbf{VOT20}} & \multicolumn{3}{c}{\textbf{VOT22}} & \multicolumn{3}{c}{\textbf{VOTS24}} & \multicolumn{3}{c}{\textbf{DiDi}} \\

\rowcolor{Gray} & \textit{Q} & \textit{Acc} & \textit{Rob} & \textit{Q} & \textit{Acc} & \textit{Rob} & \textit{Q} & \textit{Acc} & \textit{Rob} & \textit{Q} & \textit{Acc} & \textit{Rob} \\

\hline

SAM3 \cite{sam3}
& 75.5 & 80.4 & 97.3 
& 75.5 & 80.4 & 97.3 
& 72.7 & 80.4 & 87.5 
& 70.9 & 74.3 & 94.5 
\\

DAM4SAM-3
& 74.4 & 80.9 & 96.9 
& 76.6 & 81.0 & 97.4 
& 73.0 & 81.3 & 89.9 
& 72.1 & 74.7 & 96.0 
\\

SAMURAI-3
& 75.1 & 80.4 & 97.2 
& 77.4 & 80.4 & 97.7 
& 71.6 & 80.6 & 89.5 
& 71.4 & 74.3 & 95.6 
\\

SR-DRM-3
& -- & -- & -- 
& -- & -- & -- 
& -- & -- & -- 
& -- & -- & -- 
\\

SAMITE-3
& 77.2 & 80.6 & 97.8 
& 77.6 & 80.6 & 97.9 
& -- & -- & -- 
& 71.5 & 74.3 & 95.2 
\\

SA-DRM-3
& -- & -- & -- 
& -- & -- & -- 
& -- & -- & -- 
& -- & -- & -- 
\\

HiM2SAM-3
& 75.8 & 80.4 & 97.4 
& 77.3 & 80.4 & 97.7 
& 69.9 & 80.5 & 88.7 
& 71.1 & 74.7 & 94.9 
\\

HiM-DRM-3
& -- & -- & -- 
& -- & -- & -- 
& -- & -- & -- 
& -- & -- & -- 
\\

\hline

\end{tabular}
}
\end{table*}

\section{Results} \label{sec:results}
\noindent

\subsection{Quantitative Results}
\noindent We report standard tracking metrics on five large-scale OTB-style datasets (Table~\ref{tab:sam2_bbox_results} and Table~\ref{tab:sam3_bbox_results}) and four VOT-style benchmarks, including DiDi (Table~\ref{tab:sam2_vot} and Table~\ref{tab:sam3_vot}). All numbers are in \%.

\paragraph{SAM2-based baselines.}
\noindent Table~\ref{tab:sam2_bbox_results} shows that increasing SAM2 capacity yields consistent gains across datasets (e.g., LaSOT success $68.5 \!\rightarrow\! 73.6$ from SAM2-T to SAM2-L).
Replacing FIFO with DAM4SAM improves SAM2 substantially on long and distractor-heavy datasets:
DAM4SAM-L reaches \textbf{75.1/83.3/81.1} (S/NP/P) on LaSOT and \textbf{60.9/75.3/72.2} on LaSOT$_{\text{ext}}$, improving over SAM2-L by $+6.6$ (S) and $+4.1$ (S), respectively.
On GOT-10k, DAM4SAM-L attains \textbf{81.1 AO} with \textbf{91.4/77.2} SR$_{0.50}$/SR$_{0.75}$, compared to SAM2-L at 80.8 AO and 91.3/75.5.

Among SAM2-derived trackers, SAMITE and HiM2SAM are competitive with DAM4SAM on several datasets.
For instance, SAMITE-B achieves \textbf{74.9/83.4/81.4} on LaSOT (comparable to DAM4SAM-L on precision), while HiM2SAM-L achieves \textbf{61.3/75.7/72.8} on LaSOT$_{\text{ext}}$.

\noindent\textbf{Hybrid (SAM2): SR-DRM.}
SR-DRM improves over the corresponding DAM4SAM variants on the datasets where it is evaluated (Table~\ref{tab:sam2_bbox_results}).
Notably, SR-DRM-B improves DAM4SAM-B on LaSOT by \textbf{+1.3/+1.5/+1.4} (S/NP/P) reaching \textbf{74.5/82.8/80.4}, and improves DAM4SAM-B on TNL2K by \textbf{+1.4/+1.1/+1.3} reaching \textbf{59.8/79.2/65.9}.
On LaSOT$_{\text{ext}}$, SR-DRM-T improves DAM4SAM-T by \textbf{+1.5/+2.0/+1.8} reaching \textbf{58.9/72.6/69.4}.
(Other hybrids contain missing entries in the current tables and are therefore not compared here.)

\paragraph{SAM2-based VOT evaluation.}
\noindent Table~\ref{tab:sam2_vot} reports performance on VOT20/22/VOTS24 and DiDi.
Within SAM2, larger capacity generally improves Q/Acc (e.g., VOT22: SAM2-T $62.3$ Q $\rightarrow$ SAM2-L \textbf{69.2} Q).
DAM4SAM provides consistent gains over SAM2 at matched capacity on most benchmarks:
for example, VOT22 improves from SAM2-L $69.2$ Q to DAM4SAM-L \textbf{75.0} Q, and DiDi improves from SAM2-L $64.7$ Q to DAM4SAM-L \textbf{69.4} Q.
SAMITE variants are competitive on VOT-style metrics (e.g., VOT22: SAMITE-L \textbf{72.5} Q), while HiM2SAM variants show mixed behavior depending on benchmark (e.g., VOTS24: HiM2SAM-B \textbf{67.4} Q).

\paragraph{SAM3-based baselines.}
\noindent Table~\ref{tab:sam3_bbox_results} shows that SAM3 substantially improves over SAM2 across all five datasets.
For example, SAM3 achieves \textbf{74.7/82.0/78.8} on LaSOT and \textbf{63.8/78.5/75.5} on LaSOT$_{\text{ext}}$, while reaching \textbf{84.5 AO} on GOT-10k and \textbf{86.6/91.2/88.2} on TrackingNet.
SAM3Long provides small additional gains on most datasets (e.g., LaSOT precision $78.8 \rightarrow 79.6$).
Within SAM3-based methods, DAM4SAM-3 improves on LaSOT and LaSOT$_{\text{ext}}$ (e.g., LaSOT S $74.7 \rightarrow \textbf{76.5}$; LaSOT$_{\text{ext}}$ S $63.8 \rightarrow \textbf{66.0}$), while SAMITE-3 attains the strongest LaSOT numbers among the listed SAM3 baselines (\textbf{77.4/85.1/82.4}).
Hybrid SAM3 results (SR-DRM-3/SA-DRM-3/HiM-DRM-3) are currently blank in Table~\ref{tab:sam3_bbox_results} and are not included in quantitative comparisons here.

\paragraph{SAM3-based VOT evaluation.}
\noindent Table~\ref{tab:sam3_vot} shows strong VOT performance for SAM3-family trackers.
On VOT20, SAMITE-3 achieves the best reported Q among the listed methods (\textbf{77.2}).
On VOT22, DAM4SAM-3 reaches \textbf{76.6} Q, while SAMITE-3 reaches \textbf{77.6} Q.
On VOTS24, DAM4SAM-3 improves over SAM3 (Q $72.7 \rightarrow \textbf{73.0}$ and Rob $87.5 \rightarrow \textbf{89.9}$).
On DiDi, DAM4SAM-3 achieves the best reported Q/Rob among the listed methods (\textbf{72.1} Q, \textbf{96.0} Rob).
Hybrid SAM3 VOT results are currently missing in Table~\ref{tab:sam3_vot} and are omitted from comparison.

\section{Conclusion}
\noindent This work studied memory as the main driver of robustness in SAM-based visual object tracking across model generations. 
Our evaluation on a diverse suite of OTB-style and VOT-style benchmarks shows that: (i) naive FIFO updates in SAM2 are prone to error accumulation, while memory-aware designs (e.g., DAM4SAM, SAMITE, HiM2SAM) consistently improve long-term stability; and (ii) SAM3 provides a strong base improvement, but still benefits from explicit memory design beyond conservative confidence-based retention.

Guided by the modular RAM/DRM decomposition in DAM4SAM, we introduced a hybrid memory view in which the short-term RAM admission rule can be replaced by alternative selection policies from the literature while keeping DRM unchanged. 
The initial hybrid instantiation (SR-DRM) yields consistent gains over its DAM4SAM counterparts on the datasets where results are currently available, validating that short-term memory policies and long-term distractor resolution are complementary rather than redundant.

Overall, our findings suggest that progress in SAM-based tracking is increasingly bottlenecked by memory design rather than backbone capacity alone. 
A practical outcome is a unified, plug-and-play memory framework that enables controlled transfer of SAM2-era memory logic to SAM3 and supports systematic hybridization of short-term selection with long-term disambiguation. 
Completing the missing hybrid results under SAM3 and expanding the analysis to additional failure modes (e.g., extreme occlusion and identity switches) are promising next steps.